\documentclass[sigconf,nonacm]{acmart}  
\AtBeginDocument{%
  }

\setcopyright{acmlicensed}
\copyrightyear{2026}
\acmYear{2026}
\acmDOI{XXXXXXX.XXXXXXX}
\acmConference[KDD ’26]{Make sure to enter the correct
  conference title from your rights confirmation email}{August 09-13,
  2026}{Jeju, Korea}
\acmISBN{978-1-4503-XXXX-X/2018/06}

\usepackage{enumitem}       
\usepackage{multirow}       
\usepackage{array}          
\usepackage{graphicx}       
\usepackage{tabularx}       
\usepackage{tcolorbox}      
\usepackage{algorithm}
\usepackage{algorithmic}
\usepackage{hyperref}

\begin{document}

\title{Heaven-Sent or Hell-Bent? Benchmarking the Intelligence and Defectiveness of LLM Hallucinations}

\author{Chengxu Yang}
\affiliation{%
  \institution{Wuhan University of Technology}
  \institution{Hubei Key Laboratory of Transportation Internet of Things}
  \institution{BreathingCORE}
  \country{China}
}
\email{311368@whut.edu.cn}

\author{Jingling Yuan}
\authornote{Corresponding Authors}
\affiliation{%
  \institution{Wuhan University of Technology}
  \institution{Hubei Key Laboratory of Transportation Internet of Things}
  \country{China}
}
\email{yjl@whut.edu.cn}

\author{Siqi Cai}
\affiliation{%
  \institution{Wuhan University of Technology}
  \institution{Hubei Key Laboratory of Transportation Internet of Things}
  \country{China}
}
\email{csiqi@whut.edu.cn}

\author{Jiawei Jiang}
\affiliation{%
  \institution{Wuhan University}
  \country{China}
}
\email{jiawei.jiang@whu.edu.cn}

\author{Chuang Hu}
\authornotemark[1]
\affiliation{%
  \institution{Wuhan University}
  \country{China}
}
\email{chuanghu@um.edu.mo}



\renewcommand{\shortauthors}{Yang et al.}

\begin{abstract}

Hallucinations in large language models (LLMs) are commonly regarded as errors to be minimized. However, recent perspectives suggest that some hallucinations may encode creative or epistemically valuable content, a dimension that remains underquantified in current literature. Existing hallucination detection methods primarily focus on factual consistency, struggling to handle heterogeneous scientific tasks and balance creativity with accuracy. To address these challenges, we propose HIC-Bench, a novel evaluation framework that categorizes hallucinations into Intelligent Hallucinations (IH) and Defective Hallucinations (DH), enabling systematic investigation of their interplay in LLM creativity. HIC-Bench features three core characteristics: (1) Structured IH/DH Assessment. using a multi-dimensional metric matrix integrating Torrance Tests of Creative Thinking (TTCT) metrics (Originality, Feasibility, Value) with hallucination-specific dimensions (scientific plausibility, factual deviation); (2) Cross-Domain Applicability. spanning ten scientific domains with open-ended innovation tasks; and (3) Dynamic Prompt Optimization. leveraging the Dynamic Hallucination Prompt (DHP) to guide models toward creative and reliable outputs. The evaluation process employs multiple LLM judges, averaging scores to mitigate bias, with human annotators verifying IH/DH classifications. Experimental results reveal a nonlinear relationship between IH and DH, demonstrating that creativity and correctness can be jointly optimized. These insights position IH as a catalyst for creativity and reveal the ability of LLM hallucinations to drive scientific innovation.Additionally, the HIC-Bench offers a valuable platform for advancing research into the creative intelligence of LLM hallucinations.\footnote{The code and dataset are available at \href{https://github.com/chujiguangniao/HIC-bench}{https://github.com/chujiguangniao/HIC-bench}}

\end{abstract}







\maketitle

\section{Introduction}
In recent years, large language models (LLMs) have achieved remarkable progress in diverse domains, including natural language processing (NLP), complex reasoning, and scientific discovery \citep{liu2024exploring,hou2024wikicontradict}. Notably, their generative capabilities have been successfully applied to scenarios such as medical diagnosis, financial analysis, and scientific research \citep{guo2024large,xi2025rise}. However, a critical challenge that has emerged is the phenomenon of hallucinations in their generated output, which has become a significant bottleneck in their deployment, as extensively documented in the literature. Hallucinations are typically characterized as instances where the model's output deviates from factual accuracy or user expectations, posing a substantial risk in applications where reliability is paramount. To address this issue, the research community has developed several evaluation benchmarks aimed at quantifying and analyzing hallucinations, including TruthfulQA \citep{lin2021truthfulqa}, UHGEval \cite{liang2023uhgeval}, and HalluDial \citep{luo2024halludial}. These benchmarks primarily focus on assessing the truthfulness and reliability of the content generated by LLMs.


However, the essence of this phenomenon goes far beyond mere errors. Research suggests that LLM hallucinations reflect, to some extent, the core attributes of human creative cognition, specifically the capacity for divergent exploration and recombination beyond the boundaries of established knowledge \citep{surveyhallandcreat,halperin2024artificial,liu2024exploring}. This ability to transcend factual constraints bears a striking resemblance to the unconstrained imagination exhibited by humans in artistic innovation and scientific breakthroughs. For example, in the domain of protein design, "hallucinated" proteins generated by LLM structures that, although absent, were subjected to experimental validation have been demonstrated to exhibit stable configurations and functional properties in a range of tested scenarios \citep{anishchenko2021novo,hallucinationproteins}. Similarly, in the field of robotic navigation, a novel paradigm termed "Learning from Hallucinations" has been proposed \citep{Learningfromhallucination}, in which innovative patterns emerging from hallucinations are leveraged to optimize path planning for robotic systems. These findings collectively indicate that hallucinations, far from being solely a deficiency, may also represent an emergent manifestation of creative reasoning.

\begin{figure}[htbp] 
    \centering 
    \includegraphics[width=\linewidth]{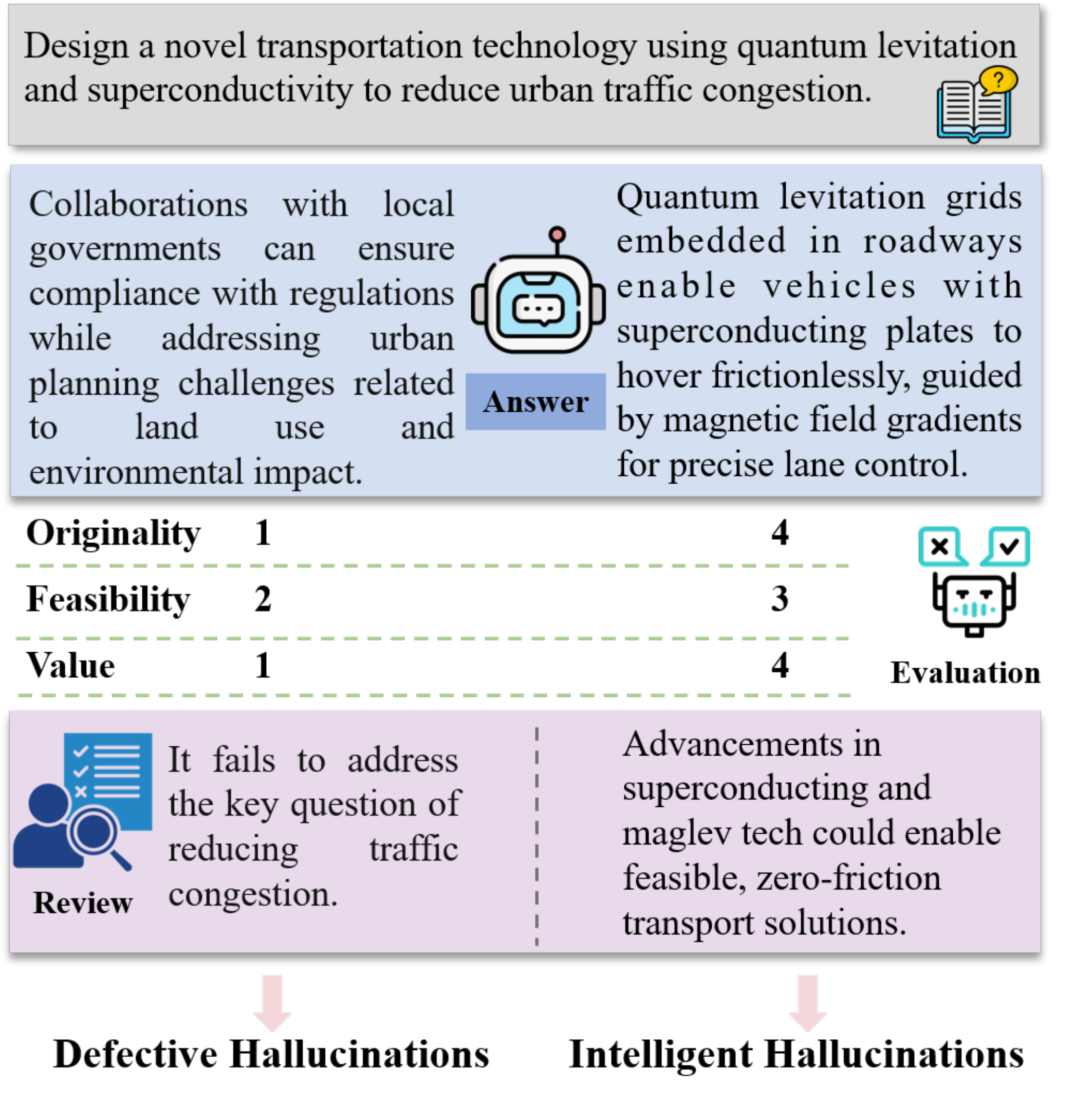} 
    \caption{Example Instances of Hallucination Types, Evaluation, and Review in HIC-Bench}
    \label{fig:p1} 
\end{figure}

Drawing on this emerging perspective, this paper investigates the relationship between LLM hallucinations and creativity. As shown in Figure \ref{fig:p1}, we propose a distinction between two categories of hallucinations: Defective Hallucinations (\textbf{DH}), which encompass content contradicting established facts or scientific principles \citep{surveyDhallucination}, and Intelligent Hallucinations (\textbf{IH}), which, though misaligned with reality, are grounded in plausible reasoning and exhibit innovative characteristics \citep{surveyIhallucination}. It is posited that IH constitutes a distinct dimension of LLMs' generative capabilities and may catalyze scientific advancements \citep{CreativityinAI}. To this end, we explore strategies to mitigate DH prevalence while preserving and enhancing IH proportion, augmenting LLMs' utility in creative tasks. This investigation departs from the conventional paradigm focused on "reducing hallucinations," offering a novel lens to elucidate parallels between LLMs and human cognition.


To substantiate this perspective, we introduce \textbf{HIC-Bench} (Hallucination \& Innovation Classification Benchmark), a novel evaluation benchmark designed to quantify the hallucination performance of LLMs in open-ended creative tasks. Unlike prior work, which predominantly focuses on eliciting induced hallucinations, HIC-Bench is informed by LLMs creativity assessment tasks \citep{Artorartifice, Enhancingthecreativity, ruan2024liveideabench} and integrates hallucination detection methodologies \citep{liang2023uhgeval} to construct an open-domain question-answering dataset spanning ten scientific disciplines. This dataset, meticulously crafted to incorporate domain-specific core principles and real-world requirements, aims to authentically emulate the reasoning processes inherent in scientific innovation. By employing varied prompt strategies, we systematically examine their impact on the prevalence of DH and IH. Experimental findings suggest that reducing DH does not necessarily precipitate a decline in innovativeness, indicating that it may be feasible to diminish DH while preserving or even enhancing the proportion of IH.


In summary, our key contributions are as follows:
\begin{itemize} 
    \item We introduce a new perspective to extract the most effective hallucination component in LLMs and define it as a valuable resource for advancing scientific innovation. 
    \item We propose HIC-Bench, a comprehensive evaluation benchmark that integrates the assessment of LLMs’ scientific creativity with hallucination analysis distinguished by intelligent and defective, incorporating a dataset spanning ten scientific domains to evaluate hallucination performance in creative tasks.
    \item We introduce the Dynamic Hallucination Prompt (DHP) pipeline, which facilitates dynamic refinement of model generation by iteratively refining the construction of positive and negative examples, enabling the mitigation of the defective nature of hallucinations while enhancing their intelligent aspect in creative scientific contexts.
    \item Evaluation results elucidate that the relationship between intelligent hallucinations and defective hallucinations does not exhibit a simple positive correlation but appears nonlinear, modulated by multiple factors, suggesting that intelligent hallucinations can be preserved or enhanced under suitable constraints to foster the development of novel scientific concepts.
\end{itemize}


\section{Related Works}
\label{relate}
Our work is inspired by studies in LLM creativity assessment, hallucination classification, and evaluation methodologies.

\subsection{Hallucinations in LLMs}
LLMs frequently generate hallucinations, outputs lacking factual grounding, a phenomenon classified by various frameworks. Huang et al. categorize hallucinations into factual and faithful types \citep{surveyDhallucination}. Li et al. separate them into intrinsic and extrinsic forms \citep{bang2025hallulens}. Zhang et al. define them as Input-Conflicting, Context-Conflicting, and Fact-Conflicting \citep{zhang2023siren}. Furthermore, benchmarks like HaluEval \citep{li2023halueval}, UHGEval \citep{liang2023uhgeval}, and HalluDial \citep{luo2024halludial} quantitatively analyze these hallucinations but often neglect their creative potential. Building on this foundation, HIC-Bench uniquely distinguishes Intelligent Hallucinations (IH), which promote creativity, from Defective Hallucinations (DH), which indicate inaccuracies, emphasizing their role in scientific innovation through a human creativity perspective.


\subsection{Assessing Creativity in LLMs}
The creative capacity of LLMs in generative artificial intelligence has lately elicited considerable scholarly interest. Prevailing methods for evaluating creativity, grounded in the Torrance Tests of Creative Thinking (TTCT) \citep{TTCT}, derive from Guilford’s framework of divergent thinking \citep{guilford1967nature}. These methodologies assess four principal dimensions: (1) originality: the aptitude for devising novel concepts; (2) elaboration: the proficiency in refining and expanding notions; (3) fluency: the capability to generate abundant ideas; (4) flexibility: the facility for reasoning across disparate domains. Extending prior investigations on LLMs' creativity \citep{Enhancingthecreativity}, we propose an evaluation system specifically calibrated for hallucination-driven creativity. Two critical challenges persist in incorporating hallucinations into this paradigm: (1) redefining metrics of innovation for LLMs in scientific endeavors to delineate IH from DH; (2) modifying and augmenting established creativity frameworks to precisely measure IH’s contributions.


\subsection{LLMs-as-a-judge}

As LLMs increasingly permeate natural language processing tasks, their functionality has expanded beyond content generation to include automated evaluation of text, code, and scientific outputs. Their capacity to assess quality in open-ended questions, creative tasks, and complex reasoning scenarios has garnered attention \citep{Llms-as-judges}. Research indicates that, in certain contexts, LLMs’ evaluations align closely with human expert judgments \citep{JudgingLLM-as-a-judge}, underscoring their potential in automated assessment. Existing frameworks for LLMs typically focus on tasks with fixed answers, such as text quality, code accuracy, and factual consistency, where performance depends on matching reference standards. Open-ended tasks, like scientific innovation evaluation or conceptual divergence testing, however, lack singular correct responses. Drawing on TTCT metrics and previous definitions of LLM’s creative capacity \citep{CreativityinAI}, we propose a refined Prompt design framework for evaluating scientific creativity. This framework establishes metrics across creativity and hallucination dimensions, while enforcing strict variable control to ensure consistency and interpretability in LLMs’ assessments.

\section{HIC-bench: From scientific hallucinations to controllable innovation}
\label{headings}

\begin{figure*}[t] 
    \centering 
    \includegraphics[width=\textwidth]{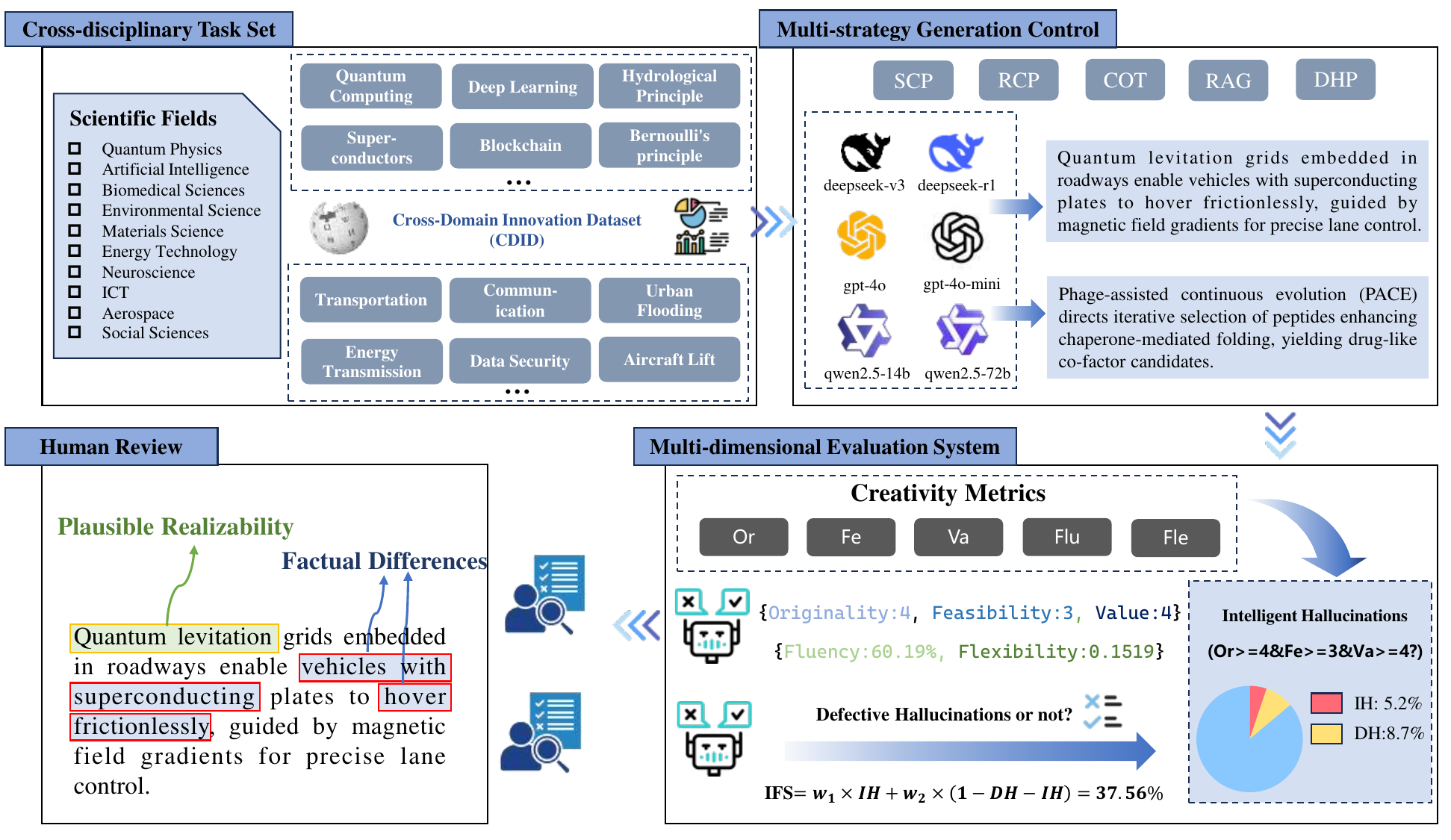} 
    \caption{Design of the HIC-Bench Framework: We first construct a cross-disciplinary dataset by integrating Wikipedia principles with real-world problems. We then evaluate the model’s responses across various generation strategies, analyzing its creativity and hallucinations through automated evaluation. Finally, human reviewers perform token-level analysis on the model’s evaluation outputs to enhance their reliability.“Or” for Originality, “Fe” for Feasibility, “Va” for Value, “Flu” for Fluency, and “Fle” for Flexibility.}
    \label{fig:p2} 
\end{figure*}

In this section, we elaborate on the design philosophy and core components of HIC-Bench, a modular and extensible benchmark designed to evaluate hallucinations of LLMs in scientific creativity tasks. Departing from traditional hallucination studies focused solely on error correction, HIC-Bench leverages cognitive mechanisms of human creative thinking to systematically explore the dual nature of LLMs’ outputs, namely DH and IH. Through a meticulously designed cross-domain task set, dynamic generation strategies, and a multi-tiered evaluation system, HIC-Bench offers researchers a structured platform to deeply investigate the role of LLMs’ hallucinations in scientific innovation. Its core workflow is illustrated in Figure \ref{fig:p2}.


\subsection{Framework Structures}



HIC-Bench addresses two pivotal research questions: Do traditional hallucination mitigation strategies inadvertently curtail LLMs’ creative capacity? Is it feasible to devise a scientific approach that diminishes DH while amplifying IH, thereby optimizing both innovation and reliability? Beyond serving as an assessment tool, HIC-Bench embodies a paradigm shift, moving from suppressing hallucinations to cultivating their beneficial forms. We combine task-driven goals with the intrinsic properties of LLM into a three-layer modular system that supports cross-model evaluation, scalable dataset integration, and highly repeatable automated workflows:


\textbf{Cross-disciplinary Task Set}. HIC-Bench constructs an open-ended task set spanning ten pivotal scientific domains, replicating the generative challenges LLMs face in real-world scientific innovation, with questions designed to embody theoretical depth, interdisciplinary complexity, and cutting-edge relevance. Unlike prior datasets assessing only knowledge breadth or factual accuracy, this approach deliberately incorporates "\textbf{fuzzy factual boundaries}", eliciting outputs that blend hypothetical, predictive, and analogical elements diverging from reality. Such a design unlocks LLMs’ latent creativity and hallucination expression. The dataset’s question formulation examines not merely knowledge retrieval proficiency, but emphasizes LLMs’ capacity for generating “useful hallucinations” within uncharted problem spaces, fostering insights that transcend conventional boundaries.


\textbf{Multi-strategy Generation Control}. A suite of Prompt strategies, including Strict Constraint Prompt (SCP), Relaxed Constraint Prompt (RCP), RAG, and CoT among others, is harnessed to explore their influence on hallucination profiles in LLMs systematically. This approach investigates the interplay between innovation and factual fidelity, adapting output characteristics to varying scientific contexts.


\textbf{Multi-dimensional Evaluation System}. Conventional hallucination detection hinges on factual consistency or semantic similarity, yet we argue that in scientific creativity tasks, certain hallucinations constitute “high-dimensional cognitive outputs”, resisting simplistic classification as errors. HIC-Bench establishes a multi-dimensional evaluation system rooted in creativity assessment frameworks, precisely identifying and quantifying IH. This system merges key TTCT metrics, such as originality, feasibility, and value, with dimensions like scientific plausibility and factual deviation from hallucination analysis, forming a specialized metric matrix tailored to adjudicate LLMs’ outputs. We introduce a composite metric, Intelligent-Fidelity Score (\textbf{IFS}), which gauges the dynamic equilibrium between creativity and factual integrity in generated results.  The evaluation employs multiple LLM judges, averaging their scores to mitigate bias, with human annotators verifying intelligent versus defective hallucination (IH vs. DH) classifications, ensuring robust and interpretable assessments. Beyond enabling quantitative recognition of IH, this framework lays an actionable foundation for constructing and modulating “valuable hallucinations” in future research. For further details, please refer to Appendix \ref{A2}


\subsection{Evaluation Metrics}

We integrate metrics from human creativity assessment tasks to devise an evaluation system tailored for LLMs. Through multi-dimensional metrics and a composite scoring formula, this system quantifies the scientific merit and innovative potential of LLMs’ outputs.


\subsubsection{LLM Creativity Assessment}
Creativity hinges on novelty, often termed originality or uniqueness, and value, encompassing utility, effectiveness, or relevance \citep{CreativityinAI,barron1955disposition,stein1953creativity}. Grounded in this conceptualization, alongside the TTCT framework and LLMs’ generative traits, we establish three creativity metrics suited for scientific tasks:


\textbf{Originality(Or)}. This metric gauges the novelty and breakthroughs of generated content within academic research, evaluating whether it introduces theoretically valuable perspectives, methodological innovations, or discoveries. Originality demands that LLMs transcend existing paradigms, yielding pioneering outputs at the frontiers of knowledge rather than mere recombinations of prior work.


\textbf{Feasibility(Fe)}. This assesses the realizability of generated content under current scientific conditions and technical frameworks, encompassing the rigor of theoretical derivations, the practicality of experimental designs, and the viability of technical pathways. For LLMs’ outputs, it examines whether proposed solutions adhere to discipline-specific standards of scientific validation.


\textbf{Value(Va)}. Value or contribution evaluates the potential impact of generated content on disciplinary advancement spanning theoretical construction methodological innovation and practical utility. High-quality academic outputs should advance critical scientific questions, forge new research paradigms, or offer valuable interdisciplinary insights.


\textbf{Fluency\&Flexibility(Flu\&Fle)}. Given that LLMs excel at generating abundant ideas and reasoning across knowledge domains \citep{hurst2024gpt,chi2024evolutionary}, Fluency and Flexibility are not prioritized in this paper. However, to maintain compliance with the TTCT standards, results for Fluency and Flexibility are still presented.


Each of the first three metrics is scored on a 5-point Likert scale, with explicit criteria defined to ensure transparency and consistency, thereby mitigating ambiguity in LLMs’ evaluation. For Fluency, similarity assessment of model outputs is conducted using SimCSE \citep{gao2021simcse}. Meanwhile, flexibility is evaluated by calculating the variance in scores across different questions.   

\subsubsection{IH and DH}
We design the evaluation of IH and DH to classify hallucinations based on the aforementioned metrics, leveraging them to distinguish the Intelligent potential and limitations of LLM outputs in a structured manner.

\textbf{IH}. Hallucinations in LLMs share parallels with human innovative thinking, producing reasoning outcomes that diverge from the status quo yet potentially harbor groundbreaking ideas for scientific advancement \citep{surveyhallandcreat}. This intelligence is described as results or novel idea combinations unlikely to emerge from most individuals \citep{surveyIhallucination}. To isolate IH from standard knowledge-based responses, we define generated content as valuable IH if it satisfies three creativity metrics, originality and value each scoring $\geq 4$, and feasibility $\geq 3$ on a 5-point scale, reflecting high innovation and worth despite partial factual divergence, while retaining plausible realizability. The IH ratio is formulated in Equation~\ref{eq:IH_ratio}, where $N_{IH}$ denotes the count of IH instances, and $N_{total}$ represents the total number of generated outputs.
\begin{equation}
    IH_{ratio}=\frac{N_{IH}}{N_{total}} \label{eq:IH_ratio}
\end{equation}


\textbf{DH}. Generated content classified as DH includes outputs marred by factual errors, logical inconsistencies, or severe violations of scientific principles, lacking innovation or practical utility. We employ LLMs for automated scientific validation of responses, supplemented by human evaluation to confirm fidelity. Consistency is further assessed using kwPrec \citep{liang2023uhgeval}, a keyword-segmentation metric. The DH ratio is expressed in Equation~\ref{eq:DH_ratio}, where $N_{DH}$ indicates the count of DH instances. 
\begin{equation}
    DH_{ratio}=\frac{N_{DH}}{N_{total}} \label{eq:DH_ratio}
\end{equation}


Notably, $IH_{ratio} + DH_{ratio} \neq 1$, as outputs neither innovative nor factually errant are classified as neutral noise responses, distinct from both IH and conventional DH traits.


\subsubsection{IFS}
To evaluate the balanced performance of LLMs across creativity and hallucination tendencies, we propose the IFS, a unified metric that integrates results from creativity assessment and hallucination classification. This score offers a comprehensive measure of the quality of generated content. The computation of IFS is detailed in Equation~\ref{eq:ifs}, where $w_1$ and $w_2$ are weight parameters summing to 1, enabling adaptability to diverse task requirements. For instance, creative writing tasks may prioritize imaginative hallucinations with a higher $w_1$, whereas medical applications demand greater accuracy with an elevated $w_2$. In our evaluation, $w_1$ and $w_2$ are set to 0.6 and 0.4 respectively, prioritizing intelligent hallucinations while ensuring output fidelity.
\begin{equation}
    IFS = w_1 \times IH_{ratio} + w_2 \times(1-DH_{ratio}-IH_{ratio}) \label{eq:ifs}
\end{equation}


\subsection{Dataset}
We construct a cross-domain dataset supporting HIC-Bench’s evaluation capabilities, spanning ten scientific fields: Quantum Physics, Artificial Intelligence, Biomedical Sciences, Environmental Science, Materials Science, Energy Technology, Neuroscience, Information and Communication Technology(ICT), Aerospace, and Social Sciences. These domains encompass a wide range from foundational science to applied technology, ensuring the assessment captures diversity and representativeness. 


\textbf{Design of Open-ended Questions and Hallucination Mechanisms}. Open-ended questions are crafted with the intent of unleashing LLMs’ creative potential while grounding their outputs in reasoned speculation and innovative reasoning drawn from established knowledge. Task descriptions explicitly require LLMs to harness domain-specific insights, such as accounting for quantum entanglement constraints in quantum communication protocols, or incorporating recent deep learning advances into disease diagnostic methods. This structure promotes novel solutions, eliciting outputs that transcend simple recombinations of existing knowledge and offer plausible conjectures not fully bound by reality.


\textbf{Cross-Domain Innovation Dataset (CDID)}. Drawing on core principles and techniques extracted from Wikipedia, we build a knowledge foundation, enriched with real-world frontier challenges like transportation and communication, shaping open-ended, cross-disciplinary innovation tasks. Assessing LLMs’ innovative diversity comprehensively, we elicit ten responses per task, analyzing 6000 responses across six models and ten domains, covering 100 distinct innovation tasks. Response generation employs diverse strategies under meticulous control to maintain variability. Additionally, we establish a specialized knowledge base dataset (CDKB) for these strategies, blending Wikipedia data with LLMs’ in-depth analyses of task-specific principles, thereby enhancing knowledge precision.


The CDID dataset contains 100 open-domain innovation tasks. This scale is consistent with contemporary practices in the evaluation of generative language models, where task quality and design often outweigh dataset size in determining benchmarking effectiveness. Empirical evidence suggests that compact yet diverse benchmarks can still offer meaningful perspectives on model behavior. For instance, HumanEval \citep{HumanEval} and Bamboogle \citep{Bamboogle}, containing 164 and 125 items respectively, have become standard references for code generation and question answering tasks. Furthermore, to ensure statistical rigor, we also conduct significance tests across all compared models. For further details, please refer to Appendix \ref{B}


\section{EXPERIMENTS \& DISCUSSION}
\label{others}

\begin{table*}[t!]
\caption{Creativity and Hallucination Performance of LLMs Across Temperature Settings (T).“Or” for Originality, “Fe” for Feasibility, “Va” for Value, “Flu” for Fluency, and “Fle” for Flexibility, Where $\uparrow$ Denotes Higher Values Are Better and $\downarrow$ Denotes Lower Values Are Better}
\label{tab:llm_comparison1}
\centering
\begin{tabular}{llcccccccc}
\toprule
\multirow{2}{*}{\textbf{LLM}} & \multirow{2}{*}{\textbf{T}} & \multicolumn{5}{c}{\textbf{Creativity}} & \multicolumn{3}{c}{\textbf{Hallucination}} \\
\cmidrule(lr){3-7} \cmidrule(lr){8-10}
& &  \textbf{Or}$\uparrow$ & \textbf{Fe}$\uparrow$ & \textbf{Va}$\uparrow$ & \textbf{Flu}$\downarrow$ & \textbf{Fle}$\downarrow$ & \textbf{IH}$\uparrow$& \textbf{DH}$\downarrow$ & \textbf{IFS}$\uparrow$ \\
\midrule
\multirow{2}{*}{deepseek-v3} & 1.0 & 2.93 & 3.82 & 3.22 & 63.00\% & 0.15 & 13.40\% & 3.20\% & 41.40\% \\
 & 0.4 & 2.90 & 3.79 & 3.18 & 62.46\% & 0.14 & 12.40\% & 1.40\% & 41.92\% \\
 \midrule
\multirow{2}{*}{gpt-4o-mini} & 1.0  & 2.62 & 3.73 & 2.98 & 63.54\% & 0.09 & 2.90\% & 5.00\% & 38.58\% \\
 & 0.4 & 2.57 & 3.74 & 2.97 & 63.74\% & 0.08 & 2.20\% & 4.50\% & 38.64\% \\
 \midrule
\multirow{2}{*}{gpt-4o} & 1.0  & 2.85 & 3.86 & 3.14 & 62.39\% & 0.14 & 9.60\% & 1.30\% & 41.40\% \\
 & 0.4  & 2.75 & 3.82 & 3.06 & 63.00\% & 0.14 & 7.10\% & 0.60\% & 41.18\% \\
\midrule
 \multirow{2}{*}{qwen2.5-14b} & 1.0  & 2.54 & 3.65 & 2.96 & 59.35\% & 0.12 & 5.20\% & 8.70\% & 37.56\% \\
 & 0.4  & 2.47 & 3.67 & 2.93 & 59.17\% & 0.13 & 4.70\% & 7.90\% & 37.40\% \\
 \midrule
\multirow{2}{*}{qwen2.5-72b} & 1.0  & 2.80 & 3.76 & 3.17 & 65.51\% & 0.11 & 8.00\% & 1.70\% & 40.92\% \\
 & 0.4  & 2.69 & 3.74 & 3.13 & 66.13\% & 0.11 & 6.60\% & 0.90\% & 40.85\% \\
\bottomrule
\end{tabular}
\end{table*}

\begin{figure*}[t!] 
    \centering 
    \includegraphics[width=\textwidth]{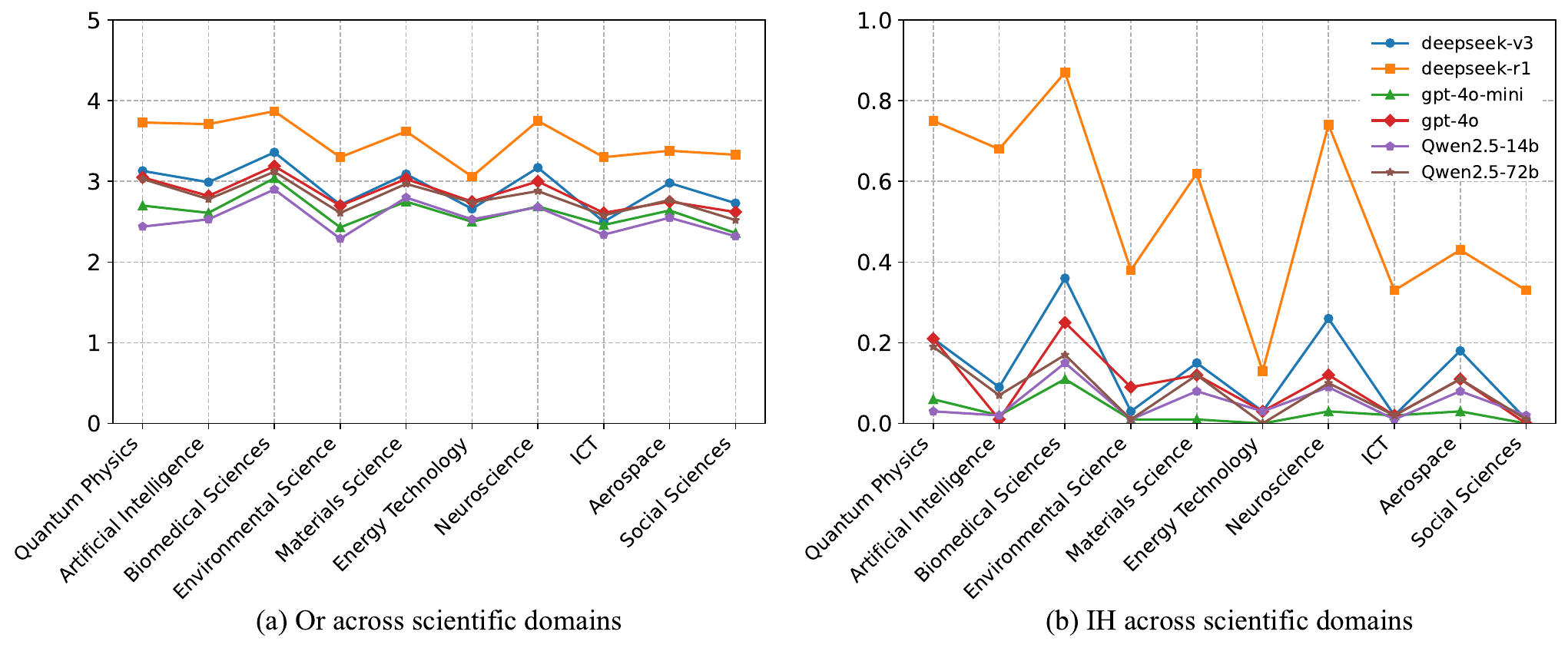} 
    \caption{Comparative Analysis of Originality and IH Across Scientific Domains}
    \label{fig:OrandIH} 
\end{figure*}

This chapter employs HIC-Bench to evaluate LLMs’ creativity and hallucination performance on the CDID dataset while concurrently exploring the influence of diverse prompt strategies on IH and DH. Experiments utilize the CDID dataset and CDKB knowledge base, encompassing two parallel investigations: the impact of hallucination mitigation techniques on preserving IH, and methods to reduce DH while maintaining IH. These studies advance the understanding of balancing factual fidelity with innovative output in scientific contexts, identifying dynamic adjustments to enhance model performance.

\textbf{Selected LLM Models}. Six technically diverse LLMs are selected to evaluate HIC-Bench’s applicability across varied architectures and optimization approaches: gpt-4o-2024-11-20 \citep{hurst2024gpt} serves as a benchmark for general-purpose performance; gpt-4o-mini \citep{achiam2023gpt} represents lightweight models optimized via knowledge distillation; qwen2.5-14b-instruct \citep{bai2023qwen} and qwen2.5-72b-instruct \citep{yang2024qwen2} enable analysis of parameter scale effects; deepseek-v3 \citep{liu2024deepseek}, refined through reinforcement learning, acts as a foundation model; and deepseek-r1 \citep{guo2025deepseek} excels in specialized CoT reasoning capabilities. This selection establishes a systematic evaluation framework, facilitating comparative analysis of model compression techniques, parameter scaling, reinforcement learning strategies, and dedicated reasoning architectures.


\textbf{Experimental Hyperparameters}. Generation employs two foundational strategies: SCP (Strict Constraint Prompt) and RCP (Relaxed Constraint Prompt). SCP enforces stringent control through prompts demanding factual adherence, while RCP fosters diversity by encouraging innovative thinking. Tuning hyperparameters across varied LLMs presents complex challenges, as models differ in architecture and respond distinctly to identical configurations. The principle of harmonizing output stability with creative latitude while preserving cross-model consistency shapes the settings. Generation temperature, set at 1.0, balances creativity with scientific rigor, producing nuanced responses for open-ended tasks. Evaluation temperature, fixed at 0, ensures precision in distinguishing IH and DH during the assessment. The maximum token length is set to 70. These configurations enable a thorough evaluation of LLMs’ hallucination and creativity performance, supporting consistent comparisons across diverse prompt strategies.


\subsection{Evaluating Creativity and Hallucination Dynamics in LLMs}

\textbf{Temperature Effects on Creativity and Hallucination}. To investigate how temperature adjustments influence LLMs’ creativity and hallucination profiles, we compare settings of 1.0 as the baseline and 0.4 for conservative outputs, using the standard SCP prompt to assess HIC-Bench under controlled variations. Deepseek-r1 \citep{guo2025deepseek} is excluded due to its lack of temperature support. Table \ref{tab:llm_comparison1} indicates that the decrease in temperature to 0.4 reduces DH in all models, with gpt-4o-mini \citep{achiam2023gpt} declining from 5.00\% to 4.50\% and showing the most significant change, though IH also decreases, with deepseek-v3 dropping from 13.40\% to 12.40\%. This behavior reflects temperature’s regulation of the model’s generation distribution, where lower temperatures lead models to favor high-probability conservative outputs, enhancing precision at the expense of creative exploration \citep{temperatureeffect}. IFS scores remain largely stable across temperatures, suggesting that temperature adjustments primarily redistribute IH and DH without altering the overall innovation-factuality balance. These findings, consistent with prior studies, affirm HIC-Bench’s capability to capture nuanced behavioral shifts in LLMs under temperature variations.

\textbf{Comparative Analysis of Originality and IH Across Scientific Domains}. We examine the generative creativity of large language models across diverse disciplinary domains by evaluating two key dimensions: Originality (Or) and Intelligent Hallucinations (IH), under SCP prompts. Figure~\ref{fig:OrandIH} (a) presents the distribution of originality scores, where most models exhibit heightened creativity in Biomedical Sciences and Aerospace, while comparatively lower scores are observed in Environmental Science, Social Sciences, and Energy Technology. This variation suggests that generative originality is not uniform but instead shaped by the epistemic structure of each domain. Figure~\ref{fig:OrandIH} (b) displays the corresponding IH proportions, which follow a broadly similar trend. Among all evaluated systems, deepseek-r1 consistently outperforms its counterparts on both metrics, reflecting its advanced capacity for creative text generation. These findings collectively highlight the domain-sensitive dynamics of LLM creativity, demonstrating systematic shifts in performance across disciplinary boundaries.


\begin{table*}[t!]
\caption{Creativity and Hallucination Performance of Six LLMs Under Diverse Prompt Strategies, with CoT and RAG Constructed on the SCP Baseline. Bold values indicate the best-performing model–strategy pairs. "P" denotes the applied prompt strategy.}
\label{tab:llm_comparison2}
\centering
\begin{tabular}{llccccccccc}
\toprule
\multirow{2}{*}{\textbf{LLM}} & \multirow{2}{*}{\textbf{P}} & & \multicolumn{5}{c}{\textbf{Creativity}} & \multicolumn{3}{c}{\textbf{Hallucination}} \\
\cmidrule(lr){4-8} \cmidrule(lr){9-11}
& & & \textbf{Or}$\uparrow$ & \textbf{Fe}$\uparrow$ & \textbf{Va}$\uparrow$ & \textbf{Flu}$\downarrow$ & \textbf{Fle}$\downarrow$ & \textbf{IH}$\uparrow$& \textbf{DH}$\downarrow$ & \textbf{IFS}$\uparrow$ \\
\midrule
\multirow{4}{*}{deepseek-v3} & SCP & &  2.93 & \textbf{3.82} & 3.22 & \textbf{63.00}\% & 0.15 & 13.40\% & 3.20\% & 41.40\% \\
 & COT & & 2.98 & 3.78 & 3.23 & 63.06\% & 0.16 & 17.00\% & \textbf{1.90}\% & 42.64\% \\
 & RAG & & 2.84 & 3.81 & 3.15 & 64.31\% & 0.14 & 9.80\% & 2.20\% & 41.08\% \\
 & RCP & & \textbf{3.13} & 3.75 & \textbf{3.33} & 63.89\% & \textbf{0.12} & \textbf{22.20}\% & 2.10\% & \textbf{43.60}\% \\
\midrule
\multirow{4}{*}{deepseek-r1} & SCP & & 3.50 & 3.75 & 3.63 & \textbf{60.19}\% & 0.15 & 52.60\% & 1.20\% & 50.04\% \\
 & COT & & 3.50 & 3.77 & 3.63 & 60.35\% & 0.17 & 52.60\% & 0.60\% & 50.28\% \\
 & RAG & & 3.46 & \textbf{3.79} & 3.59 & 60.99\% & 0.16 & 49.00\% & \textbf{0.20}\% & 49.72\% \\
 & RCP & & \textbf{3.70} & 3.57 & \textbf{3.76} & 60.46\% & \textbf{0.09} & \textbf{71.10}\% & 0.30\% & \textbf{54.10}\% \\
\midrule
\multirow{4}{*}{gpt-4o-mini} & SCP & & 2.62 & 3.73 & 2.98 & 63.54\% & 0.09 & 2.90\% & 5.00\% & 38.58\% \\
 & COT & & 2.59 & \textbf{3.75} & 2.97 & \textbf{62.89}\% & 0.09 & 2.80\% & 4.60\% & 38.72\% \\
 & RAG & & 2.58 & 3.73 & 2.95 & 64.65\% & 0.08 & 1.70\% & 3.50\% & 38.94\% \\
 & RCP & & \textbf{2.88} & 3.71 & \textbf{3.11} & 64.00\% & \textbf{0.07} & \textbf{8.30}\% & \textbf{2.60}\% & \textbf{40.62}\% \\
\midrule
\multirow{4}{*}{gpt-4o} & SCP & & 2.85 & \textbf{3.86} & 3.14 & 62.39\% & 0.14 & 9.60\% & 1.30\% & 41.40\% \\
 & COT & & 2.81 & 3.84 & 3.13 & \textbf{62.16}\% & 0.15 & 9.30\% & 0.40\% & 41.70\% \\
 & RAG & & 2.81 & 3.83 & 3.11 & 64.50\% & 0.14 & 7.90\% & \textbf{0.10}\% & 41.54\% \\
 & RCP & & \textbf{3.08} & 3.73 & \textbf{3.22} & 62.58\% & \textbf{0.10} & \textbf{18.60}\% & 0.50\% & \textbf{43.52}\% \\
 \midrule
\multirow{4}{*}{qwen2.5-14b} & SCP & & 2.54 & 3.65 & 2.96 & 59.35\% & 0.12 & 5.20\% & 8.70\% & 37.56\% \\
 & COT & & 2.50 & 3.65 & 2.93 & \textbf{58.88}\% & 0.12 & 4.30\% & 5.20\% & 38.78\% \\
 & RAG & & 2.52 & \textbf{3.68} & 2.93 & 62.79\% & \textbf{0.09} & 4.70\% & 5.80\% & 38.62\% \\
 & RCP & & \textbf{2.67} & 3.56 & \textbf{2.98} & 59.52\% & 0.12 & \textbf{6.30}\% & \textbf{4.50}\% & \textbf{39.46}\% \\
\midrule
\multirow{4}{*}{qwen2.5-72b} & SCP & & 2.80 & 3.76 & 3.17 & 65.51\% & 0.11 & 8.00\% & 1.70\% & 40.92\% \\
 & COT & & 2.77 & \textbf{3.80} & 3.14 & 65.55\% & 0.12 & 7.00\% & 1.40\% & 40.84\% \\
 & RAG & & 2.69 & 3.79 & 3.13 & 66.71\% & 0.11 & 4.80\% & 1.30\% & 40.44\% \\
 & RCP & & \textbf{2.87} & 3.65 & \textbf{3.19} & \textbf{63.27}\% & \textbf{0.10} & \textbf{10.40}\% & \textbf{1.20}\% & \textbf{41.60}\% \\
\bottomrule
\end{tabular}
\end{table*}

\subsection{Prompt Strategy Effects on Model Performance}

Reducing hallucinations in LLMs without compromising efficacy is challenging \citep{lee2023mathematical}, as they include DH and a creative facet as IH. As such, reducing DH while preserving IH is a key research goal. However, overly strict constraints may suppress creative capabilities \citep{creativityunderconstraints}. Table \ref{tab:llm_comparison2} compares the SCP, CoT, RAG, and RCP strategies’ impact on IH, DH and IFS. More details are provided in the Appendix\ref{F3}.

 

\textbf{SCP}. As the base strategy, SCP enforces logical coherence and creativity through structured prompts designed to integrate existing research insights. It aims to balance innovation with factuality, fostering IH generation while maintaining factual integrity in scientific contexts. This approach provides a foundational framework for evaluating hallucination profiles across different models. Results guide HIC Bench’s hallucination assessment and support further strategy comparisons.


\textbf{CoT}. CoT uses “Let’s think step by step” to boost accuracy \citep{zero-shotcot} via incremental reasoning. It lowers DH and raises IFS but may reduce IH in some models like gpt-4o. Model response varies with this approach.


\textbf{RAG}. RAG integrates the CDKB dataset for factual grounding, effectively reducing DH. However, this approach also constrains IH, revealing a trade-off between reliability and creative output. highlighting a trade-off between reliability and creativity in open-ended scientific tasks.


\textbf{RCP}. RCP relaxes SCP constraints to prioritize innovation and value, boosting IH across models. Unexpectedly, it also reduces DH, suggesting moderated constraints enhance creativity while lowering factual deviations.


\subsection{Sensitivity Analysis of IFS}
\begin{figure}[htbp] 
    \centering 
    \includegraphics[width=\linewidth]{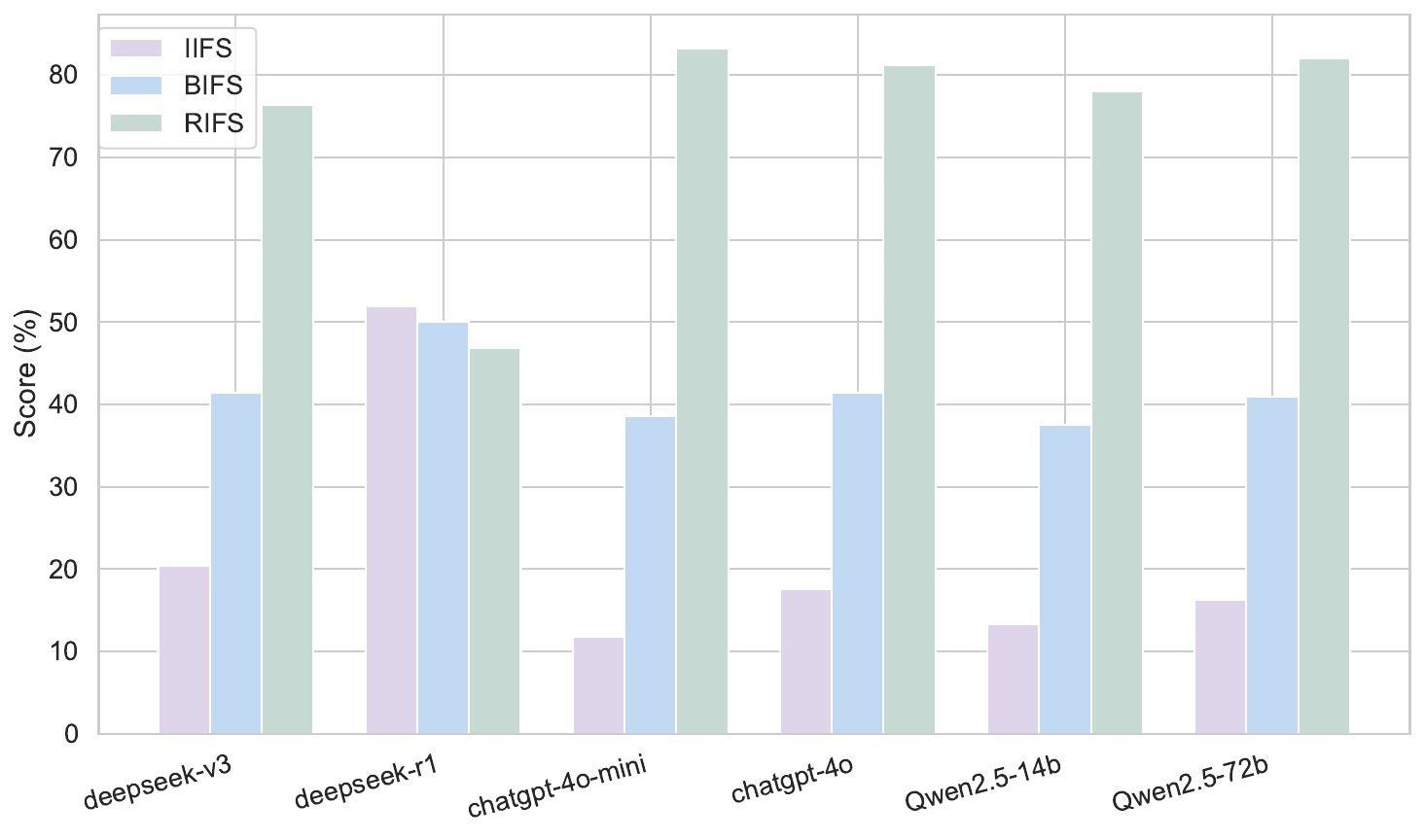} 
    \caption{Comparative analysis of model indicators across scientific fields under SCP}
    \label{fig:IFS} 
\end{figure}

We present a horizontal comparison of model performance under different IFS scoring scenarios in the HIC-Bench evaluation. Figure \ref{fig:IFS} visualizes the results for Intelligent IFS (IIFS, $ w_1 = 0.9 $) applied in scenarios emphasizing intelligent hallucinations, Balanced IFS (BIFS, $ w_1 = 0.6 $) used in balanced creativity and fidelity contexts, and Robust IFS (RIFS, $ w_1 = 0.1 $) utilized in scenarios requiring high accuracy. In the first two metrics IIFS and BIFS deepseek r1 achieves higher scores demonstrating strong performance in creativity driven scenarios. However in accuracy focused environments as measured by RIFS this model is less suitable due to challenges in maintaining precision. 

These findings highlight the trade offs in model performance across different scenarios. Models excelling in creativity driven contexts may struggle in precision focused tasks emphasizing the need to select appropriate models based on the specific requirements of each application.

\subsection{Dynamic Prompt Effects on Hallucination Profiles}


To address the challenge of mitigating DH while enhancing IH, we propose a novel pipeline, Dynamic Hallucination Prompt (DHP). The DHP pipeline optimizes response generation and evaluation in HIC-Bench by dynamically adapting prompts based on real-time feedback. Initially, human experts input positive examples—responses scoring high on Torrance Tests of Creative Thinking (TTCT) metrics and negative examples—responses with high DH rates to establish a baseline. During machine evaluation, responses with IFS scores exceeding the initial positive examples are set as new positive examples, while negative examples are continuously updated with the latest DH responses. This observation motivated the design of the Dynamic Hallucination Prompt (DHP), which provides examples to guide model outputs toward more distinct and reliable responses. Detailed procedures are provided in Appendix \ref{D}.


Ablation experiments on gpt-4o and gpt-4o-mini use SCP as the baseline, with results shown in Table~\ref{tab:llm_hallucination3}. With DHP positive prompting, DH drops notably to 0.90\% for gpt-4o-mini and 0.10\% for gpt-4o, while IH increases moderately. When DHP is combined with RCP constraints, IH further rises to 11.10\% and 21.10\%, with DH reduced to 1.70\% and 0.30\%, respectively. This setting also yields the highest IFS scores at 41.54\% for gpt-4o-mini and 44.10\% for gpt-4o, exceeding their SCP baselines. These outcomes underscore the efficacy of DHP in balancing innovation and factuality, outperforming prior strategies in creative generation tasks.



\begin{table}[h]
\caption{Ablation Study on Dynamic Hallucination Prompt (DHP) and RCP with SCP as Baseline.}
\label{tab:llm_hallucination3}
\centering
\begin{tabular}{lcccccc}
\toprule
\textbf{LLM} & \textbf{RCP} & \textbf{DHP} & \textbf{IH}$\uparrow$ & \textbf{DH}$\downarrow$ & \textbf{IFS}$\uparrow$ \\
\midrule
\multirow{4}{*}{gpt-4o-mini} 
&  &  & 2.90\% & 5.00\% & 38.58\% \\
& \checkmark &  & 8.30\% & 2.60\% & 40.62\% \\
&  & \checkmark & 5.50\% & 0.90\% & 40.74\% \\
& \checkmark & \checkmark & 11.10\% & 1.70\% & 41.54\% \\
\midrule
\multirow{4}{*}{gpt-4o} 
&  &  & 9.60\% & 1.30\% & 41.40\% \\
& \checkmark &  & 18.60\% & 0.50\% & 43.52\% \\
&  & \checkmark & 12.40\% & 0.10\% & 42.44\% \\
& \checkmark & \checkmark & 21.10\% & 0.30\% & 44.10\% \\
\bottomrule
\end{tabular}
\end{table}




\section{Limitations and Future Work}

HIC-Bench has advanced the understanding of IH and DH interplay in LLMs and the evaluation of mitigation strategies, yet limitations persist. The framework primarily focuses on structured tasks like question-answering, neglecting areas such as literary writing. Future work will extend to broader creativity tasks for a comprehensive assessment. Additionally, balancing IH and DH remains challenging due to potential biases from inductive prompts and the need for refined IH metrics. Future efforts will develop domain-specific prompts and integrate human-involved multi-dimensional evaluations to enhance IH assessment accuracy and manage IH-DH dynamics. Lastly, the framework will be applied to multimodal and cross-lingual benchmarks to validate its generalizability, while examining the ethical implications of promoting IH.


\section{Conclusion}
This paper introduces HIC-Bench, a benchmark tailored to assess the interplay between Intelligent Hallucinations (IH) and Defective Hallucinations (DH) in large language models (LLMs). The framework’s reliability is validated through temperature parameter analysis, with results elucidating the hallucination profiles of mainstream LLMs in creative tasks, followed by a comparative evaluation of four strategies to mitigate hallucinations while minimizing their adverse impact on IH. Notably, our findings reveal that the relationship between IH and DH is not simply positively correlated, suggesting that it may be feasible to enhance creative potential while reducing DH. To this end, the Dynamic Hallucination Prompt (DHP) pipeline is introduced, substantially augmenting the intelligent aspects of model hallucinations. Ultimately, this research quantifies the potential of hallucinations as a scientific dream machine, thereby paving new avenues for future hallucination studies.


\section{AI Ethics}

Although our framework distinguishes between Intelligent Hallucinations (IH) and Defective Hallucinations (DH), users must select an appropriate IFS based on the specific context, as shown in Equation \ref{eq:ifs}. For instance, in high accuracy scenarios, increasing the weight of $w_2$ is recommended, while in high innovation scenarios, elevating $w_1$ is more suitable. Furthermore, while IH may contribute to scientific advancement, it remains a form of hallucination. Users should avoid conflating IH outputs with factual content, as this could lead to misinterpreting IH as reality.

\bibliographystyle{ACM-Reference-Format}
\bibliography{sample-base}

\newpage
\clearpage
\appendix


\section{Prompt Details} 
\subsection{Multi-strategy Generation Control}
\label{A1}


\begin{tcolorbox}[colback=gray!30, colframe=gray!30] 
        \textbf{SCP (Strict Constraint Prompt)}\newline
        Assume you are an expert in \{field\}.\newline
        Please provide an answer to the following questions, not exceeding 70 tokens. Requirements:\newline
        1: Ensure feasibility by grounding in current scientific principles and technological trends;\newline
        2: Propose novel concepts or methods, avoiding unsupported speculation;\newline
        3: Ensure that the programme has value for the target area;\newline
        4: Maintain logical rigor without contradictions or vague statements.\newline
        Format: Plain text, no numbering or Markdown. \newline
        Question: \{question\}
\end{tcolorbox}

\begin{tcolorbox}[colback=gray!30, colframe=gray!30] 
\textbf{RCP (Role-Constrained Prompt)}\newline
Assume you are an expert in \{field\}.\newline
Please provide an answer to the following questions, not exceeding 70 tokens. Requirements:\newline
1: Propose novel concepts or methods, avoiding unsupported speculation;\newline
2: Ensure that the programme has value for the target area;\newline
Format: Plain text, no numbering or Markdown. \newline
Question: \{question\}
\end{tcolorbox}

\begin{tcolorbox}[colback=gray!30, colframe=gray!30] 
\textbf{COT (Chain-of-Thought Prompt)}\newline
Assume you are an expert in \{field\}.\newline
Please provide an answer to the following questions, not exceeding 70 tokens. Requirements:\newline
1: Ensure feasibility by grounding in current scientific principles and technological trends;\newline
2: Propose novel concepts or methods, avoiding unsupported speculation;\newline
3: Ensure that the programme has value for the target area;\newline
4: Maintain logical rigor without contradictions or vague statements.\newline
5: Please think step by step before answering the question.\newline
Format: Plain text, no numbering or Markdown. \newline
Question: \{question\}
\end{tcolorbox}

\begin{tcolorbox}[colback=gray!30, colframe=gray!30] 
\textbf{RAG (Retrieval-Augmented Generation Prompt)}\newline
Assume you are an expert in \{field\}.\newline
Please provide an answer to the following questions, not exceeding 70 tokens. Requirements:\newline
1: Ensure feasibility by grounding in current scientific principles and technological trends;\newline
2: Propose novel concepts or methods, avoiding unsupported speculation;\newline
3: Ensure that the programme has value for the target area;\newline
4: Maintain logical rigor without contradictions or vague statements.\newline
5: Answers should primarily rely on the provided Wikipedia principles; if information is insufficient, general scientific knowledge may be used, but speculation and contradictions must be avoided:\{principle\}.\newline
Format: Plain text, no numbering or Markdown. \newline
Question: \{question\}
\end{tcolorbox}

\subsection{Evaluation Prompt}
\label{A2}

For our automated evaluation, we utilized \textbf{gpt-4o} and \textbf{deepseek-v3 models}, adopting a separate assessment approach. This method reduces the bias toward self-generated responses within the same conversation, ensuring impartial evaluation. Additionally, using two distinct models mitigates the risk of single model bias. Below is our evaluation prompt, which provides detailed explanations for each indicator.

\begin{tcolorbox}[colback=gray!30, colframe=gray!30] 
\textbf{System Prompt}\newline
You are a rigorous evaluator specialized in assessing the scientific innovation quality of LLM-generated answers.
Be highly critical and avoid giving high scores to generic or vague answers.\newline
1. Evaluation Criteria (1-5 Scale):\newline
- Originality:
1 = Common knowledge or widely known concepts with no novelty.
2 = Slight variation of known methods, minimal innovation.
3 = Moderate originality, combining existing ideas in a non-trivial way.
4 = Highly novel approach, limited prior research but plausible.
5 = Breakthrough idea with no direct precedent."
If the answer only combines existing concepts without a novel mechanism, the originality score must not exceed 3.
\newline
- Feasibility:
1 = Violates fundamental scientific principles.
2 = Theoretical possibility but lacks experimental validation.
3 = Feasible under specific assumptions but challenging to implement.
4 = Technically viable with existing technology.
5 = Readily implementable with minor adaptations of existing techniques.\newline
- Value:
1 = No practical use or improvement over existing methods.
2 = Minor improvement, limited impact."
3 = Meaningful contribution with clear advantages.
4 = Significant improvement over current methods.
5 = Potential revolutionary impact on the field.
If the answer lacks technical details or fails to demonstrate security improvements, the value score must not exceed 3. \newline
2. Hallucination Detection:"
- If any of the following conditions are met, mark 'Hallucination: Yes':
- The answer does not align with the core requirements of the question.
- The answer deviates from reality.
- The answer contradicts established scientific principles.
- The answer provides irrelevant or tangential information without addressing the problem.
- The answer contains false information or made-up claims.\newline
3. Scoring Rules:
 Generic or vague responses must receive lower scores: 'Originality <= 3' \& 'Value <= 3'.
 Ensure a clear distinction between general answers and true innovations—avoid inflated scores. \newline
4. Output format:
'Originality: [1-5] Feasibility: [1-5] Value: [1-5] Hallucination: Yes/No'.
\end{tcolorbox}

\begin{tcolorbox}[colback=gray!30, colframe=gray!30] 
\textbf{User Prompt}\newline
[User Questions]:\{question\}\newline
[Answers to be evaluated]:\{answer\}
\end{tcolorbox}

\section{HIC-bench Dataset}
\label{B}
The Cross Domain Innovation Dataset (CDID) uses open ended tasks for creativity evaluation as datasets with definitive answers are not suitable. Assessing creativity involves examining the novelty feasibility and value of ideas which are often subjective and context dependent. Open ended tasks allow models to generate diverse responses enabling a more comprehensive evaluation of their ability to produce innovative and interdisciplinary solutions in scientific contexts.

In each experimental setting, we evaluate 1,000 generated responses per task. This design ensures that, despite the relatively compact number of tasks, the dataset retains strong statistical coverage and response diversity. At the same time, it supports a reliable assessment of model creativity and hallucination patterns, capturing nuanced variations within and across domains. The compact yet dense structure of CDID thus balances breadth with analytical depth, offering a practical and rigorous basis for evaluation.

\subsection{Cross-Domain Innovation Dataset (CDID)}
\label{B1}
This section introduces the CDID used in HIC-Bench. Table \ref{tab:cdid_examples} provides a subset of examples from CDID, covering domains such as Quantum Physics and Social Sciences. The dataset is an open-ended question-answering collection constructed by integrating Wikipedia’s knowledge organization principles with real-world complex problems. Core concepts from Wikipedia across interdisciplinary domains are combined with practical challenges to form innovative, open-ended questions.

\begin{table*}[h] 
\centering 
\caption{Examples from the CDID Dataset: Open-Ended Questions Across Multiple Domains} 
\label{tab:cdid_examples} 
\begin{tabular}{p{4cm}p{3cm}p{5cm}} 
\toprule
\textbf{Domain} & \textbf{Principle\&Challenge} & \textbf{Question} \\ 
\midrule 
Quantum Physics & quantum levitation, traffic & Design a novel transportation technology using quantum levitation and superconductivity to reduce urban traffic congestion. \\ 
\midrule 
Artificial Intelligence & Graph Neural Networks, social network & Propose a novel social network analysis tool based on Graph Neural Networks (GNN). \\ 
\midrule 
Biomedical Sciences & tissue engineering, artificial skin fabrication &  Propose a novel artificial skin fabrication method based on tissue engineering.\\ 
\midrule 
Environmental Science & hydrological principles, urban flooding & Propose an urban stormwater management system based on hydrological principles to mitigate urban flooding. \\
\midrule 
Materials Science & biomimetic materials, bulletproof vest & Propose a novel bulletproof vest that enhances protection and comfort, inspired by the structural characteristics of biomimetic materials. \\
\midrule 
Energy Technology & ocean energy utilization technology, electricity & Design a tidal power generation device based on ocean energy utilization technology to supply electricity to coastal areas. \\
\midrule 
Neuroscience&  principles of synaptic plasticity, enhance learning abilities & Propose an educational method to enhance learning abilities based on the principles of synaptic plasticity. \\
\midrule 
Information and Communication Technology(ICT)& optical fiber communication principles, cloud computing & Propose a novel ultra-high-speed data transmission system based on optical fiber communication principles to optimize cloud computing center networks. \\
\midrule 
Aerospace & Bernoulli's principle, aircraft lift & Design a novel wing to enhance aircraft lift based on Bernoulli's principle. \\
\midrule 
Social Sciences & social network analysis, prevent cybercrimes & Propose a strategy to prevent cybercrimes and ensure cybersecurity based on social network analysis. \\
\bottomrule
\end{tabular} 
\end{table*}

\subsection{Cross-Domain Knowledge Base Dataset (CDKB)}
\label{B2}

This section introduces the Cross-Domain Knowledge Base Dataset (CDKB) developed for HIC-Bench. CDKB is a specialized knowledge base dataset designed to provide concise yet precise domain knowledge for diverse innovation tasks. For CDKB, we first used an LLM to condense Wikipedia entries, keeping only the core definitions of domain-specific terms and stripping away extraneous text. This ensures the RAG context focuses solely on the target concept without distracting or overly verbose material that could dilute the model’s attention or introduce noise.  By integrating CDKB into Retrieval-Augmented Generation (RAG) prompts, we enable LLMs to access structured, focused, and principle-aware knowledge, thereby gaining a clearer and more accurate understanding of the underlying concepts relevant to each task. This integration ensures that the generated responses are not only creative but also well-grounded in reliable domain-specific knowledge. Table \ref{tab:cdkb_examples} showcases examples from CDKB.

\subsection{Significance Testing}
\label{B3}
Given that our tasks are open-ended rather than fixed-response, they must elicit genuinely creative answers; otherwise they would be meaningless for this study. Following recent practice (e.g., HumanEval), we prioritize quality over quantity. We first harvest trending scientific keywords and then expand each into a question plus an associated knowledge base. Every task is designed to solicit highly creative responses, and the multiple answers generated satisfy the fluency requirement of the TTCT. To rule out randomness, we perform significance tests on the IH ratio across models and strategies. Specifically, we apply a two-sided paired permutation test: we take the vector of task-level IH-proportion differences d between two models, randomly flip the sign of each element, and recompute the mean difference 10 000 times under the null hypothesis that the expected difference is zero, thereby constructing the null distribution. The two-tailed p-value was derived from the extremity of the observed mean difference within this distribution. We report the standard error, and the p-value to quantify both magnitude and statistical significance. Most results pass the 1\% significance level, and the remainder pass the 5\% level. Table \ref{tab:scp} summarizes the pairwise significance tests between models under the SCP setting.

\begin{table*}[h] 
\centering 
\caption{Examples from the CDKB Dataset: Knowledge Base for Cross-Disciplinary Innovation Tasks} \label{tab:cdkb_examples} 
\begin{tabular}{p{4cm}p{8cm}} 
\hline 
\textbf{Domain} & \textbf{Knowledge} \\ 
\hline 
Quantum Physics & Quantum levitation, also known as quantum locking, occurs when a superconductor is cooled below its critical temperature and expels magnetic fields from its interior (Meissner effect), allowing it to lock in space above a magnet due to flux pinning.  \\ 
\hline 
Artificial Intelligence & A graph neural network (GNN) is a class of artificial neural networks for processing data that can be represented as graphs.\\ 
\hline 
Biomedical Sciences &  Gene editing is a type of genetic engineering in which DNA is inserted, deleted, modified, or replaced in the genome of a living organism.\\ 
\hline 
Environmental Science &  Hydrological principles study the distribution, movement, and properties of water on Earth, encompassing the water cycle processes such as precipitation, evaporation, infiltration, and runoff.\\ 
\hline 
Materials Science & Biomimetic materials are synthetic materials designed to imitate the properties and functions of natural biological materials, often resulting in enhanced performance. \\ 
\hline 
Energy Technology & Ocean energy utilization encompasses technologies that capture energy from oceanic sources, such as tidal movements, converting it into electricity. \\ 
\hline 
Neuroscience &  Synaptic plasticity refers to the ability of synapses, the connections between neurons, to strengthen or weaken over time in response to increases or decreases in their activity, playing a crucial role in learning and memory.\\ 
\hline 
Information and Communication Technology(ICT) & Optical fiber communication uses light signals transmitted through fiber-optic cables to achieve high-speed data transmission over long distances, essential for optimizing networks in cloud computing centers. \\ 
\hline 
Aerospace & Bernoulli's principle states that an increase in the speed of a fluid occurs simultaneously with a decrease in pressure, which is fundamental in understanding how airfoil shapes generate lift in aircraft.\\ 
\hline 
Social Sciences & Social network analysis examines the relationships and interactions within a network, aiming to understand how these connections influence individual and group behaviors.  \\ 
\hline 
\end{tabular} 
\end{table*}

\begin{table*}[t]
\centering
\caption{Significance Tests Between Models under SCP. Values indicate standard errors. *** : p < 0.01, ** : p < 0.05, * : p < 0.1}
\label{tab:scp}
\begin{tabular}{lcccccc}
\toprule
 & \textbf{gpt-4o} & \textbf{gpt-4o-mini} & \textbf{qwen2.5-14b} & \textbf{qwen2.5-72b} & \textbf{deepseek-v3} & \textbf{deepseek-r1} \\
\midrule
gpt-4o        & ---   & 0.016*** & 0.014*** & 0.015**  & 0.016**  & 0.030*** \\
gpt-4o-mini   &       & ---      & 0.009**  & 0.011*** & 0.018*** & 0.033*** \\
qwen2.5-14b   &       &          & ---      & 0.012**  & 0.017*** & 0.033*** \\
qwen2.5-72b   &       &          &          & ---      & 0.017*** & 0.032*** \\
deepseek-v3   &       &          &          &          & ---      & 0.030*** \\
deepseek-r1   &       &          &          &          &          & ---      \\
\bottomrule
\end{tabular}

\end{table*}

\section{Human Review}
\label{C}
We integrate human review into HIC-Bench to enhance evaluation rigor. Specifically, we conduct a human review on the model’s Intelligent Hallucinations (IH) and Defective Hallucinations (DH) outputs, assessing their factuality and feasibility from multiple perspectives. This process ensures a more reliable distinction between intelligent and defective hallucinations in scientific contexts, strengthening the overall assessment of LLMs’ creative potential.

Due to the low efficiency and high labor cost of human review as well as the difficulty of evaluating open ended tasks automated evaluation metrics are becoming mainstream. However to ensure the reliability of automated evaluations we conducted a human review of the model classified IH and DH. This section presents examples from the human review process for model outputs classified as IH and DH. Human reviewers perform token level analysis to determine the extent to which the outputs align with reality focusing on the accuracy of concepts and terminology. For IH reviewers identify the hallucinated portions that are not fully realistic while assessing the reasonable and feasible aspects that contribute to their creative value. For DH reviewers evaluate whether the response addresses the given problem and check for obvious factual errors that render the output unrealistic or irrelevant. Additionally reviewers assess the feasibility of the proposed ideas to ensure reliable categorization. Through token level analysis we evaluated the rationality of the classifications. Table \ref{tab:human_review} provides examples of the human review process.

\begin{table*}[htbp] 
\centering 
\caption{Illustrative Examples of Hallucination with Review Insights} \label{tab:human_review} 
\begin{tabular}{p{7cm}p{9cm}} 
\hline 
\textbf{Answer} & \textbf{Review} \\ 
\hline 
Quantum levitation grids embedded in roadways enable vehicles with superconducting plates to hover frictionlessly, guided by magnetic field gradients for precise lane control.  & This response exhibits Intelligent Hallucination as it proposes a creative solution with some feasibility—quantum levitation and magnetic field gradients are grounded in physics. However, frictionless hovering and widespread superconducting vehicles are not fully realistic due to current technological and energy constraints, yet the idea holds innovative value for future transportation systems.  \\ 
\hline 
Stretchable graphene-MXene heterostructure supercapacitors woven into textiles provide high-energy-density storage with rapid charge cycles for self-powered wearables.  & This response shows Intelligent Hallucination with an innovative idea. Graphene MXene heterostructures and textile integration are feasible. However high energy density storage and rapid charge cycles in stretchable textiles for self powered wearables are not fully realistic yet. The concept holds value for future wearable energy solutions.\\ 
\hline 
A dual acting compound modulating GABA A positive allosteric modulation and NMDA receptor antagonism to balance inhibitory excitatory signaling during sleep wake transitions. &  This response shows Intelligent Hallucination with a novel concept. Modulating GABA A and NMDA receptors to balance signaling is theoretically feasible. However achieving precise control during sleep wake transitions is not fully realistic with current pharmacology. The idea offers innovative value for future neurological treatments.\\ 
\hline 
Adopt a hemispherical resonator gyroscope array with machine learning based noise filtering enabling sub arcsecond attitude determination for closed loop control stability. &  This response shows Intelligent Hallucination with an innovative approach. Hemispherical resonator gyroscopes and machine learning noise filtering are feasible. However achieving sub arcsecond attitude determination for closed loop stability is not fully realistic with current technology. The concept holds value for future precision navigation systems.\\ 
\hline 
Collaborations with local governments can ensure compliance with regulations while addressing urban planning challenges related to land use and environmental impact. & This response is classified as Defective Hallucination. It completely fails to address the problem of designing a novel transportation technology using quantum levitation and superconductivity. Instead it focuses on government collaboration and urban planning which are unrelated to the task and lack technological feasibility in this context. \\ 
\hline 
Integrating emotion recognition systems could facilitate more nuanced interactions between humans and robots. Adapting behavior based on detected emotional states may improve cooperation safety and overall user experience during collaborative tasks. & This response is classified as Defective Hallucination. It does not address the problem of designing a novel robot control system based on reinforcement learning for autonomous decision making. Instead, it focuses on emotion recognition, which is unrelated to the task and lacks relevance to the specified technical approach. \\ 
\hline  
\end{tabular} 
\end{table*}

In the Artificial Intelligence domain, we conducted a human evaluation of 600 model responses generated under the SCP strategy, where expert reviewers manually scored and classified the outputs into Intelligent Hallucinations (IH) and Defective Hallucinations (DH) based on criteria such as factual accuracy, creative value, and feasibility. This process involved a detailed analysis of each response, comparing the human assessments with the scores automatically generated by the models to evaluate the reliability and consistency of the automated evaluation system. The results are presented in Table \ref{tab:comparison_scores}.We calculated the precision and recall of LLMs in classifying hallucinations across the 600 responses. For IH, LLMs achieved a precision of 85.4\% and a recall of 87.4\%, while for DH, the precision was 95.7\% and the recall was 88.0\%. Despite implementing strict prompts to enforce factual accuracy, LLMs show some leniency in assessing certain problems, leading to some bias in their scoring. Nevertheless, these evaluations generally reflect the variation trends of IH and DH across different prompts, providing valuable insights into hallucination dynamics.

\section{ Dynamic Hallucination Prompt (DHP)}
\label{D}
In our Fluency evaluation, we calculated the similarity of model responses within the same question. Here, we compute the similarity of responses across different questions as well as the similarity among responses identified as IH and those identified as DH in the SCP. The results are presented in Table \ref{tab:simcseinIHandDH}. The similarity of IH and DH outputs generally exceeds the similarity of answers across questions. This observation motivated the design of the Dynamic Hallucination Prompt (DHP), which provides examples to guide model outputs toward more distinct and reliable responses.

The DHP pipeline is designed to enhance the generation and evaluation of model responses in the HIC-Bench framework by dynamically adapting prompts based on real-time feedback. DHP incorporates iterative prompt refinement by identifying positive examples (high-scoring responses based on Originality, Feasibility, and Value) and negative examples (responses with hallucinations) from each batch of generated answers. This process ensures that subsequent responses are guided toward higher creativity and lower defect rates, while evaluations are saved for further analysis, balancing innovation with reliability across diverse fields.

\begin{tcolorbox}[colback=gray!30, colframe=gray!30] 
\textbf{DHP (Dynamic Hallucination Prompt)}\newline
Assume you are an expert in {field}.\newline
Please provide an answer to the following questions, not exceeding 70 tokens. Requirements:\newline
1: Ensure feasibility by grounding in current scientific principles and technological trends;\newline
2: Propose novel concepts or methods, avoiding unsupported speculation;\newline
3: Ensure that the programme has value for the target area;\newline
4: Maintain logical rigor without contradictions or vague statements.\newline
Format: Plain text, no numbering or Markdown. \newline
\{positive\_prompt\_examples\} \newline
\{negative\_prompt\_examples\} \newline
Question: \{question\}
\end{tcolorbox}

\begin{table*}[htbp]
\centering
\caption{Comparison of Human and AI Evaluations in the Artificial Intelligence Domain}
\begin{tabular}{lccccc}
\hline
\textbf{Evaluator} & \textbf{Originality} & \textbf{Feasibility} & \textbf{Value} & \textbf{IH} & \textbf{DH} \\
\hline
Human & 2.87 & 3.85 & 3.04 & 13.0\% & 4.2\% \\
LLMs & 2.91 & 3.83 & 3.11 & 14.8\% & 3.8\% \\
\hline
\end{tabular}

\label{tab:comparison_scores}
\end{table*}

\begin{table*}[htbp] 
\centering 
\caption{Similarity Metrics for IH, DH, and All Answers in SCP. All Answer represents the similarity of responses across different questions.} \label{tab:simcseinIHandDH} 
\begin{tabular}{lccc} 
\hline 
\textbf{LLM} & \textbf{All Answers} & \textbf{IH} & \textbf{DH}\\ 
\hline 
deepseek-v3 & 35.99\% & 42.10\% & 40.26\%\\
deepseek-r1 & 38.97\% & 41.38\% & 37.35\%\\
gpt-4o-mini & 37.91\% & 48.50\% & 43.90\%\\
gpt-4o & 32.67\% & 40.83\% & 35.47\%\\
qwen2.5-14b & 34.55\% & 39.82\% & 36.76\%\\
qwen2.5-72b & 35.67\% & 42.03\% & 39.91\%\\
\hline

\end{tabular} 
\end{table*}

\section{Model Selection for Evaluation}
LLMs demonstrate formidable proficiency in generating scientific text, yet their outputs frequently exhibit hallucinations, deviations from factual or logical coherence. Conventionally, such hallucinations are deemed errors to be eradicated. This paper, however, posits that hallucinations are not uniformly detrimental; certain instances, though diverging from reality, manifest remarkable innovation and foresight, harboring potential value in scientific contexts. To explore this duality, we introduce HIC-Bench, a systematic evaluation framework tailored to analyze LLMs’ generative behavior in scientific innovation tasks. For the first time, it models hallucinations’ dual attributes: DH, marked by factual inaccuracies, and IH, characterized by novel, scientifically plausible insights.

Six technically diverse LLMs are selected to evaluate HIC-Bench’s applicability across varied architectures and optimization approaches: gpt-4o-2024-11-20 \citep{hurst2024gpt} serves as a benchmark for general-purpose performance; gpt-4o-mini \citep{achiam2023gpt} represents lightweight models optimized via knowledge distillation; qwen2.5-14b-instruct \citep{bai2023qwen} and qwen2.5-72b-instruct \citep{yang2024qwen2} enable analysis of parameter scale effects; deepseek-v3 \citep{liu2024deepseek}, refined through reinforcement learning, acts as a foundation model; and deepseek-r1 \citep{guo2025deepseek} excels in specialized CoT reasoning capabilities. This selection establishes a systematic evaluation framework, facilitating comparative analysis of model compression techniques, parameter scaling, reinforcement learning strategies, and dedicated reasoning architectures.


\section{Additional Experimental Results and Analysis}
\label{F}
\subsection{Comparative Analysis of Model Indicators Across Scientific Fields}
\begin{figure*}[htbp] 
    \centering 
    \includegraphics[width=1.0\textwidth]{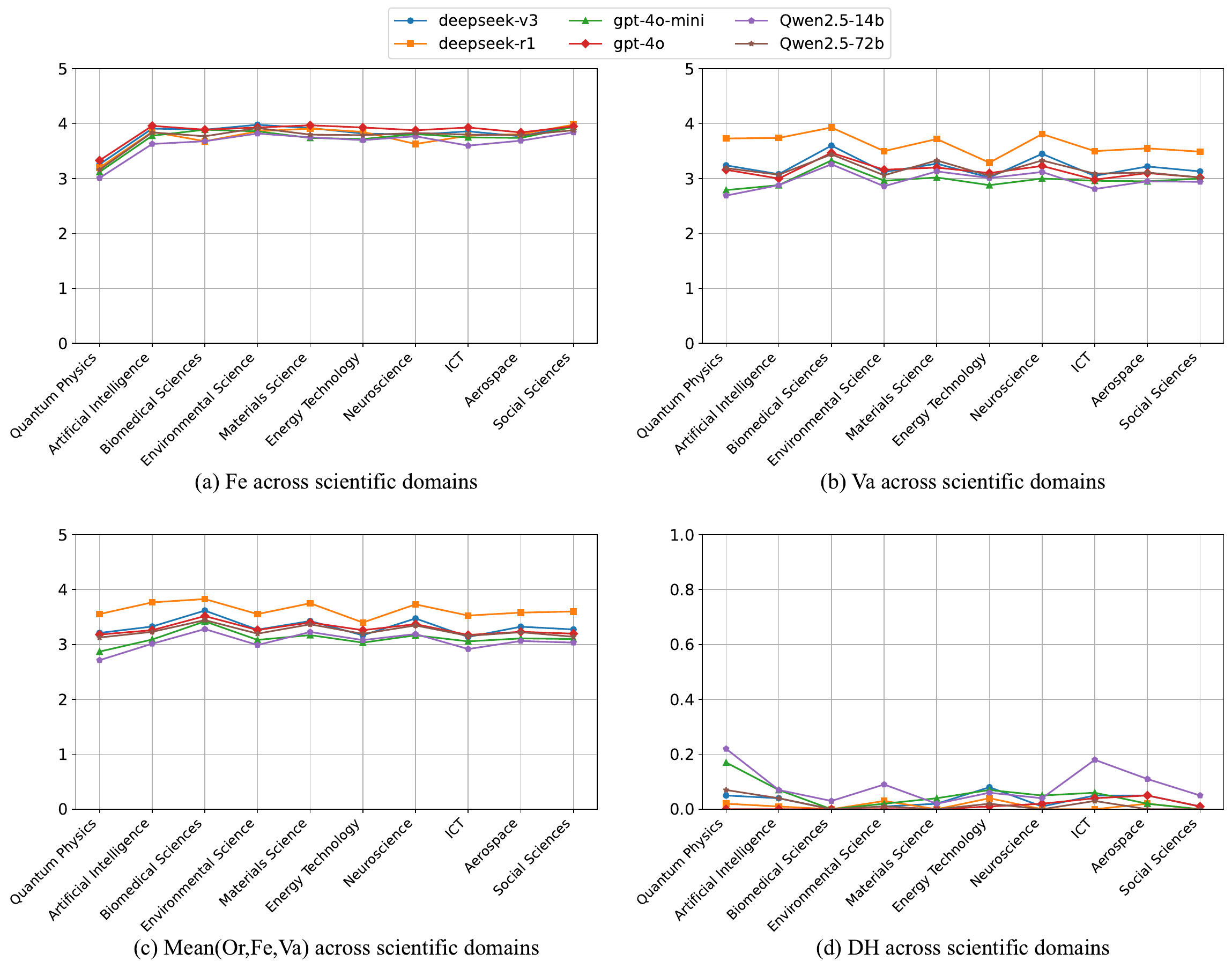} 
    \caption{Comparative analysis of model indicators across scientific fields under SCP}
    \label{fig:Or_Fe_Va_DH} 
\end{figure*}

This subsection analyzes the cross domain performance of other key indicators in the HIC-Bench evaluation by examining additional metrics across various scientific fields. Figure \ref{fig:Or_Fe_Va_DH} provides a detailed comparative analysis focusing on four indicators: Feasibility (Fe) in Subfigure (a) Value (Va) in Subfigure (b) Mean of Originality Feasibility and Value (Mean[Or, Fe, Va]) in Subfigure (c) and DH rates in Subfigure (d). Each Subfigure visualizes performance variations across domains. Feasibility remains relatively consistent across all models indicating a general tendency to produce feasible solutions. In terms of Value and contribution the deepseek-r1 outperforms others showing stronger performance in Biomedical Science and Neuroscience but lower performance in Environmental Science and Energy Technology. The mean indicator aligns with this trend.For Defective Hallucination (DH) rates Qwen2.5-14b exhibits a notably higher occurrence compared to other models. This is particularly evident in Quantum Physics and Information and Communication Technology (ICT) where the model generates responses with significant factual inaccuracies or irrelevance to the task reflecting a higher propensity for defective outputs in these complex domains.

The statistical analysis reveals substantial differences in model responses across domains. When leveraging IH for innovative generation domain applicability must be carefully considered. For instance fields like Biomedical Science require a low hallucination rate to ensure reliability. In such cases the deepseek-r1 with a higher IH rate may not be suitable despite its strong creativity as the risk of inaccurate outputs could undermine practical utility.

\begin{algorithm}[H]
\caption{Dynamic Hallucination Prompt (DHP)}
\label{alg:dhpa}
\begin{algorithmic}[1] 
\REQUIRE Question file, evaluation file, output Excel
\ENSURE Refined prompts, answer generations, evaluations

\STATE Load all questions and principles from Excel
\STATE Initialize \texttt{positive\_example} and \texttt{negative\_example} to empty

\FORALL{field $\in$ fields}
    \FORALL{question $\in$ field}
        \STATE Build prompt using current \texttt{positive\_example} and \texttt{negative\_example}
        \STATE Generate answers using the LLM
        \STATE Save generated answers
        \FORALL{answer $\in$ answers}
            \STATE Evaluate answer using LLM to get Originality, Feasibility, Value, Hallucination
            \IF{Originality $\geq 4$ \AND Feasibility $\geq 3$ \AND Value $\geq 4$}
                \IF{total score $>$ current best}
                    \STATE Update \texttt{positive\_example}
                \ENDIF
            \ENDIF
            \IF{Hallucination = Yes}
                \STATE Update \texttt{negative\_example}
            \ENDIF
        \ENDFOR
    \ENDFOR
\ENDFOR
\end{algorithmic}
\end{algorithm}


\subsection{Prompt Strategy and performance}
\label{F3}
Reducing hallucinations in LLMs without compromising their efficacy remains a formidable challenge, given that prior investigations \citep{lee2023mathematical} have elucidated, through rigorous mathematical analysis, the intrinsic interplay wherein hallucinations not only manifest as defective deviations from factual accuracy (DH) but also constitute an integral facet of model creativity, specifically IH. Consequently, identifying strategies that mitigate DH while preserving or enhancing IH emerges as a pivotal research endeavor. Although numerous methodologies have been developed to enhance the reliability of generated content, it has been posited \citep{creativityunderconstraints} that excessively stringent constraints may suppress model performance in tasks necessitating creativity. Table \ref{tab:llm_comparison2} illustrates the outcomes across varying prompt strategies, through which this section examines the influence exerted by four distinct hallucination mitigation strategies, namely SCP, CoT, RAG, and RCP, on model performance, with particular emphasis on the proportions of IH and DH and the IFS.

\textbf{SCP}: Serving as the foundational strategy, SCP enforces rigorous constraints through prompts that mandate the generation of logically coherent outputs devoid of fabricated content, while simultaneously integrating insights from existing research to foster creative responses. Within this experimental framework, prompts are meticulously crafted to enhance the generation of creative outputs, thereby facilitating a comprehensive exploration of their characteristics. Results reported in Table \ref{tab:llm_comparison2}, evaluated on the Vectara HHEM benchmark \citep{hhem-2.1-open}, reveal that deepseek-r1 exhibits a notably elevated overall hallucination rate, achieving an IH of 52.60\% and an IFS of 50.04\% in our benchmark, which surpasses other models such as gpt-4o-mini with an IH of 2.90\% and an IFS of 38.58\%. This outcome suggests that SCP effectively balances innovation with factuality to a considerable extent. In contrast, gpt-4o-mini, representing distilled architectures, demonstrates a markedly higher DH of 5.00\% under identical conditions, indicating a pronounced propensity for generating factually inaccurate content when subjected to stringent constraints.


\textbf{CoT}: By incorporating the instruction “Let’s think step by step,” the CoT strategy substantially enhances response accuracy \citep{zero-shotcot}. This zero-shot approach, which guides models through incremental reasoning processes, optimizes final outputs by generating intermediate inferential steps. Experimental findings indicate that CoT significantly elevates IFS scores while concurrently reducing DH across all models. For instance, deepseek-v3 exhibits a decline in DH from 3.20\% under SCP to 1.90\% under CoT, with its IFS rising from 41.40\% to 42.64\%. However, its impact on IH appears to diverge: while the deepseek series, such as deepseek-v3, shows an improved IH from 13.40\% to 17.00\%, suggesting that structured reasoning augments performance in creative tasks, the gpt-4o series, such as gpt-4o, experiences a decline in IH from 9.60\% to 9.30\%, potentially attributable to excessive reasoning steps constraining creative expression. Such variability underscores the differential responsiveness of models to the CoT strategy.


\textbf{RAG}: To enhance factual grounding, this strategy integrates the CDKB dataset, providing external, domain-specific knowledge and ensuring that model outputs are anchored in established scientific principles. Compared to the less constrained SCP setting, RAG imposes stronger factual guidance. Results show that RAG significantly reduces defect hallucinations. For example, gpt-4o-mini’s DH drops from 5.00\% to 3.50\%, and deepseek-v3’s decreases from 3.20\% to 2.20\%, demonstrating RAG’s effectiveness in improving factual reliability through systematic knowledge referencing. However, this improvement comes at the cost of reduced intelligent hallucinations. Deepseek-r1’s IH, for instance, falls from 52.60\% to 49.00\%, leading to a decreased in the IFS score to 49.72\%. These findings suggest that while RAG enhances factual accuracy, its strong external constraints may suppress innovative potential, highlighting a trade-off between reliability and creativity in open-ended scientific tasks.


\textbf{RCP}: Acknowledging that stringent constraints might inhibit IH, we propose the RCP strategy, which relaxes the feasibility and output restrictions of SCP, prioritizing solely the innovation and value of generated content. Results reveal that RCP significantly elevates IH across all models, while reducing Defective Hallucinations (DH) outputs in certain cases. For example, gpt-4o’s IH rises from 9.60\% under SCP to 18.60\% under RCP, yet its DH decreases from 1.30\% to 0.50\%. Similarly, qwen2.5-72b-instruct \citep{yang2024qwen2} exhibits an IH increase from 8.00\% to 10.40\%, accompanied by a DH reduction from 1.70\% to 1.20\%. This observation indicates that DH and IH are not inherently positively correlated; rather, an appropriately moderated relaxation of constraints appears to unleash the creative potential of models, while also lowering factual deviations in some models.

\subsection{Comparative Overview of LLM Hallucination Evaluation Studies}

This subsection presents Table \ref{tab:hallucination_eval_comparison} which provides a comparative overview of recent LLM hallucination evaluation studies. The table contrasts HIC-Bench with other benchmarks by examining dataset sources task types and evaluation metrics. In HIC-Bench we categorize hallucinations into Intelligent Hallucinations and Defective Hallucinations to break the traditional perspective of treating hallucinations solely as errors to be mitigated thereby offering a novel viewpoint for hallucination research. Furthermore HIC-Bench focuses on open ended cross domain innovation using real world challenges and employs creativity focused metrics such as Originality Feasibility and Value. This comparison highlights the distinct approach of HIC-Bench in evaluating LLMs for scientific creativity.

\begin{table*}[t]
\centering
\caption{Comparison of LLM Hallucination Evaluation Studies}
\label{tab:hallucination_eval_comparison}
\begin{tabular}{p{3cm}p{3cm}p{3cm}p{3cm}}
\hline
\textbf{Benchmark} & \textbf{Dataset Source} & \textbf{Task Type} & \textbf{Evaluation Metrics} \\ \hline
HHEM \citep{hhem-2.1-open} & Wiki, News & Question Answering & BLEU, ROUGE, Factuality Score \\ \hline
TruthfulQA \citep{lin2021truthfulqa} & Manual & Question Answering & Accuracy by Human or GPT Judge \\ \hline
HalluDial \citep{luo2024halludial} & Dialogues, Synthetic & Dialogue Generation & Hallucination Rate, Semantic Consistency \\ \hline
HalDetect \citep{gunjal2024detecting}& Wiki, Synthetic & Text Generation & Hallucination Rate, F1, Precision \\ \hline
FactCheck \citep{wang2024factcheck} & News, Wiki & Fact Checking & FactScore, NE Error \\ \hline
SelfCheck \citep{manakul2023selfcheckgpt} & Wiki & Fact Checking & SelfCheckGPT Score, Hallucination Rate \\ \hline
FactScore \citep{min2023factscore}& Wiki & Text Generation & FactScore, NE Error \\ \hline
HalluQA \citep{cheng2023evaluating} & Manual, Wiki & Question Answering & Non Hallucination Rate \\ \hline
UHGEval \citep{liang2023uhgeval}& News & Open Ended Generation & Accuracy, kwPrec, BERTScore \\ \hline
HIC-Bench (Ours) & Wiki, Real World Challenges & Open Ended Cross Domain Innovation & Originality, Feasibility, Value, IH, DH, IFS \\ \hline
\end{tabular}
\end{table*}


\section{Limitations and Future Work}

Due to the review process’s scope constraints, this section outlines key limitations and future directions. HIC-Bench has advanced the understanding of IH and DH interplay in LLMs and the evaluation of mitigation strategies, yet limitations persist that warrant further exploration.

\textbf{Structured Task Focus}: The framework primarily focuses on structured tasks like question answering, neglecting areas such as literary writing. Future work will extend to broader creativity tasks for a comprehensive assessment.

\textbf{Human Evaluation Constraints}: The current human evaluation only involves reviewing model outputs, without conducting a full accuracy analysis comparing human expert assessments to LLM evaluations or aligning with human preferences. Future efforts will incorporate expert evaluations and preference alignment studies to improve the robustness of IH and DH assessments.

Additionally, balancing IH and DH remains challenging due to potential biases from inductive prompts and the need for refined IH metrics. Future efforts will develop domain-specific prompts and integrate human-involved multi-dimensional evaluations to enhance IH assessment accuracy and manage IH-DH dynamics. Lastly, the framework will be applied to multimodal and cross-lingual benchmarks to validate its generalizability, while examining the ethical implications of promoting IH.


\end{document}